\pdfoutput=1
\documentclass[11pt,a4paper]{article}
\usepackage[hyperref,final]{naacl2021}
\usepackage{times}          
\usepackage{latexsym}       
\usepackage{booktabs}       
\usepackage{multirow}       
\usepackage{amsmath}        
\usepackage{amssymb}        
\usepackage{graphicx}       

\usepackage[utf8]{inputenc} 
\usepackage[T1]{fontenc}    

\usepackage{url}            
\usepackage{booktabs}       
\usepackage{amsfonts}       
\usepackage{nicefrac}       
\usepackage{microtype}      

\usepackage{latexsym}       
\usepackage{booktabs}       
\usepackage{multirow}       
\usepackage{amsmath}        
\usepackage{amssymb}        
\usepackage{graphicx}       
\usepackage{float}          

\usepackage{natbib}

\usepackage{cleveref}
\crefformat{section}{\S#2#1#3}
\crefformat{subsection}{\S#2#1#3}
\crefformat{subsubsection}{\S#2#1#3}
\crefrangeformat{section}{\S\S#3#1#4 to~#5#2#6}
\crefmultiformat{section}{\S\S#2#1#3}{ and~#2#1#3}{, #2#1#3}{ and~#2#1#3}
\Crefformat{figure}{#2Fig.~#1#3}
\Crefmultiformat{figure}{Figs.~#2#1#3}{ and~#2#1#3}{, #2#1#3}{ and~#2#1#3}
\Crefformat{table}{#2Tab.~#1#3}
\Crefmultiformat{table}{Tabs.~#2#1#3}{ and~#2#1#3}{, #2#1#3}{ and~#2#1#3}
\Crefformat{equation}{#2Eq.~(#1#3)}

\Crefformat{appendix}{#2App.~\S#1#3}

\usepackage{colortbl}
\newcommand{\CC}[1]{\cellcolor{blue!#1}}

\newcommand{\CCG}[1]{\cellcolor{green!#1}}

\definecolor{Gray}{gray}{0.92}
\definecolor{racing-green}{rgb}{0.0, 0.8, 0.6}
\definecolor{awesome-red}{rgb}{1.0, 0.13, 0.32}
\definecolor{LightCyan}{rgb}{0.88,1,1}
\definecolor{darkgreen}{RGB}{0,150,0}

\newcommand{\blu}[1][1]{\cellcolor{blue!4}}
\newcommand{\yel}[1][1]{\cellcolor{yellow!4}}
\newcommand{\gre}[1][1]{\cellcolor{green!4}}
\newcommand{\gra}[1][1]{\cellcolor{gray!4}} 

\usepackage{arydshln}
\usepackage{subcaption}

\usepackage{microtype}

\usepackage{pifont}

\usepackage{overpic}
\usepackage{rotating}
\usepackage{siunitx}
\usepackage{array}

\newcommand{\stitle}[1]{\vspace{1.8ex} \noindent{\bf #1}}

\title{Search-Optimized Quantization in Biomedical Ontology Alignment}

\author{Oussama Bouaggad$^{\ding{49},\ding{50}}$, Natalia Grabar$^{\ding{49}}$ \\
$^\ding{49}$CNRS, Univ. Lille, UMR 8163 - STL - Savoirs Textes Langage, F-59000 Lille, France \\
  $^\ding{50}$Univ. Lille, UMR 9189 - CRIStAL - Centre de Recherche en Informatique Signal \\ et Automatique de Lille, F-59000 Lille, France \\
    \texttt{\{first.last\}@univ-lille.fr}\\
}

\date{}

\begin{document}
\maketitle


\begin{abstract}
\vspace{1.25em}

In the fast-moving world of AI, as organizations and researchers develop more advanced models, they face challenges due to their sheer size and computational demands. Deploying such models on edge devices or in resource-constrained environments adds further challenges related to energy consumption, memory usage and latency. To address these challenges, emerging trends are shaping the future of efficient model optimization techniques.
From this premise, by employing supervised state-of-the-art transformer-based models, this research introduces a systematic method for ontology alignment, grounded in cosine-based semantic similarity between a biomedical layman vocabulary and the Unified Medical Language System (\textsc{UMLS}) Metathesaurus. It leverages \textsc{Microsoft Olive} to search for target optimizations among different Execution Providers (EPs) using the \textsc{ONNX Runtime} backend, followed by an assembled process of dynamic quantization employing \textsc{Intel Neural Compressor} and \textsc{IPEX} (Intel Extension for PyTorch). 
Through our optimization process, we conduct extensive assessments on the two tasks from the DEFT 2020 Evaluation Campaign, achieving a new state-of-the-art in both. We retain performance metrics intact, while attaining an average inference speed-up of 20x and reducing memory usage by approximately 70\%.\footnote{The code is available at \url{https://github.com/OussamaBouaggad/Quantization}.}


\vspace{0.5em}

\vspace{-\lastskip}
\end{abstract}
\vspace{-2.2mm}
\section{Introduction}\label{sec:intro}
\begin{figure}
    \includegraphics[width=\linewidth]{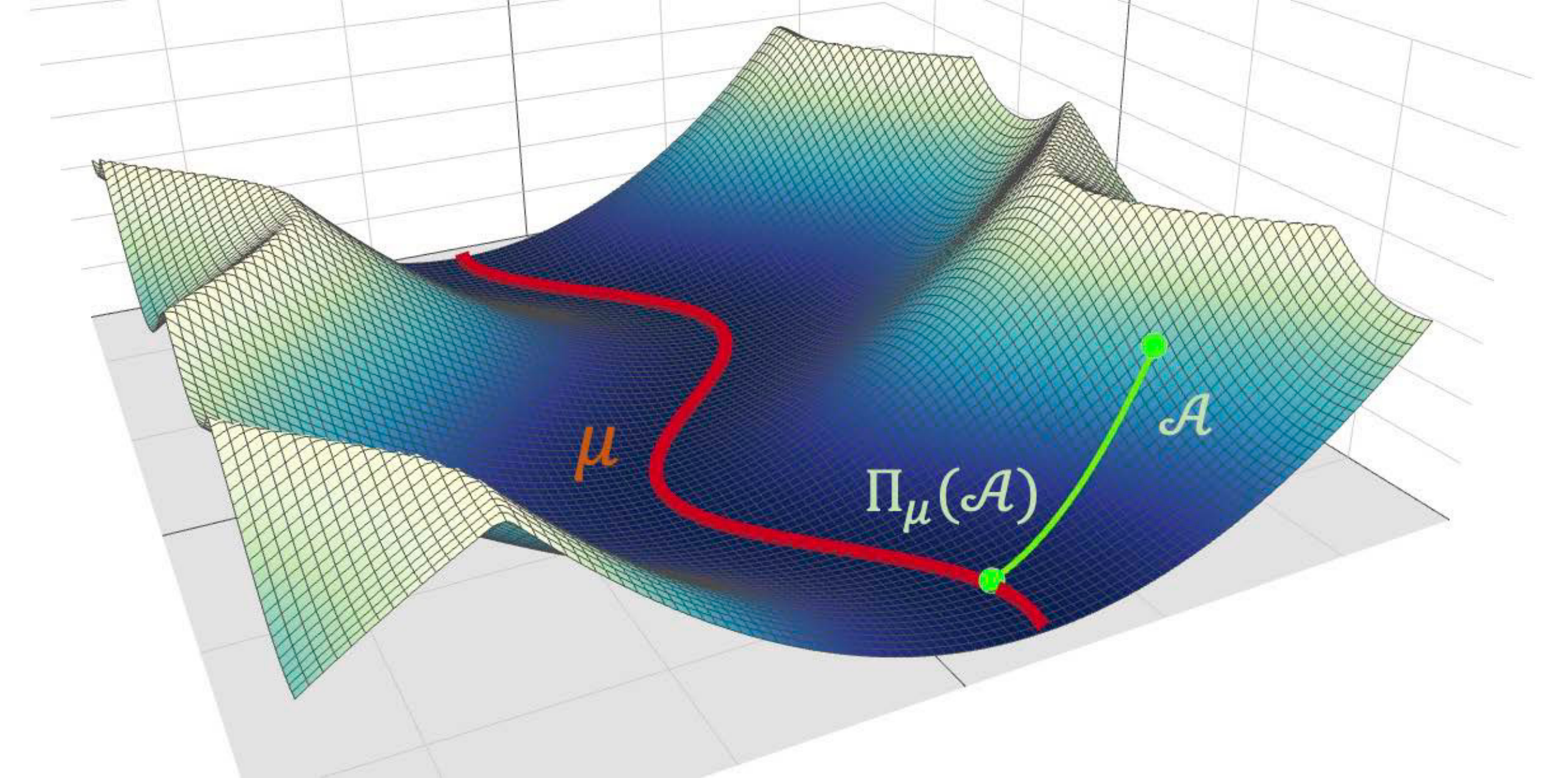}
    \caption{Starting from the initial state vector \( \mu \), the dynamic optimization trajectory (red path) guides the model toward the optimized state \( \Pi_{\mu}(A) \), monitoring the inverse spectral norm of the Hessian \( \| H^{-1} \| \) via Cholesky decomposition to achieve substantial reductions in memory usage and inference time, while preserving performance. The surface represents the loss landscape \( \mathcal{L}(\theta) \), originally used to illustrate local convexity \citep{b3fdb8de83be4269b9fa50472008efaa}, here reinterpreted to depict quantization-aware optimization, which minimizes computational overhead \( \delta \mathcal{C} \).}
    \label{fig:front}
    \vspace{-0.83em}
\end{figure}

\vspace{1.4mm}

Biomedical ontology alignment refers to the process of matching semantically related entities across diverse knowledge sources (databases) to facilitate the integration of heterogeneous data.
The historical impetus for biomedical ontology alignment arose from the need to consolidate independently developed knowledge sources, each characterized by distinct data vocabularies.
In this domain, the Unified Medical Language System (UMLS) Metathesaurus \citep{bodenreider2004unified}, developed under the auspices of the U.S. National Library of Medicine (NLM), serves as a cornerstone.\footnote{The official UMLS resource is accessible at \url{https://www.nlm.nih.gov/research/umls/index.html}.} The UMLS Metathesaurus, which comprises the most extensive collection of biomedical ontologies, including terminologies, controlled vocabularies, thesauri, and classifications, provides an essential framework for unifying standardized knowledge sources. With the ongoing evolution of this project, its size has reached over 10 million atoms, derived from more than 200 controlled vocabularies grouped into approximately 4 million concepts. Its maintenance is costly, time-consuming, and demands significant expert effort. However, decades of meticulous manual curation offer valuable material for modern supervised learning applications, establishing UMLS as a foundational resource for ontology alignment.
Conversely, the biomedical layman vocabulary \citep{koptient:hal-03095275} is designed to support the adaptation and simplification of medical texts, enhancing the accessibility of health-related documents for non-expert audiences, such as patients. Its size is steadily increasing, although it remains significantly smaller than that of large-scale terminologies. Aligning the layman vocabulary with UMLS is important for ensuring that structured medical knowledge is accessible to non-experts, thereby improving the effectiveness of healthcare communication. This helps bridge the language gap between clinicians and patients, allowing for dynamic adjustment of linguistic complexity. Nevertheless, aligning layman and expert terms accurately is challenging due to lexical variation, contextual ambiguity, and the absence of direct one-to-one mappings. Furthermore, layman expressions often lack the ontological grounding and semantic precision of formal vocabularies, making purely symbolic or rule-based methods inadequate.

Advances in Natural Language Processing (NLP), such as entity linking and semantic similarity, increasingly rely on transformer-based supervised deep learning models with specialized domain feature engineering. In this work, we propose using two approaches, the \textsc{KRISSBERT} (Knowledge-RIch Self-Supervision) model developed by Microsoft Research \citep{zhang2022knowledgerichselfsupervisionbiomedicalentity} and the large variant of the \textsc{SapBERT} model from Cambridge LTL \citep{liu2021selfalignmentpretrainingbiomedicalentity}, to align the layman vocabulary with UMLS via cosine-based semantic similarity.
The resulting biomedical alignments are manually verified by expert human annotators using a six-point rating scale (0 to 5) to assess degrees of similarity \citep{44288666a29942a9bafddedbb6c32c1e}. Additional semantic information is incorporated by including all Metathesaurus data file domains and their hierarchical structures, systematically aligned by means of a left join propagation based on the common \emph{CUI (Concept Unique Identifier)} field.

In conjunction with this, model selection is based on the distinct characteristics of each model, as no single transformer consistently handles all nuanced details and noise in alignments. Hence, a dual-model approach is used, ensuring that inaccuracies from one model are mitigated by the other. To operationalize this complementarity, alignments are merged iteratively in descending order of rating: starting with all alignments rated 5 by one model, followed by those rated 5 by the other model that are not already included, and proceeding through lower-rated alignments until a comprehensive, high-confidence set is constructed. This dualism leverages the complementary strengths of \textsc{KRISSBERT} and \textsc{SapBERT}, ensuring robust performance across diverse biomedical vocabulary contexts.
The \textsc{KRISSBERT} model addresses ambiguity and context-ignorance, particularly where entities share similar surface forms, by harnessing contextual information to improve identification accuracy. It trains a contextual mention encoder via contrastive learning with a transformer-based encoder \citep{NIPS2017_3f5ee243}, and improves linking accuracy by re-ranking top $K$ candidates with a cross-attention encoder \citep{logeswaran-etal-2019-zero,wu-etal-2020-scalable}. On the other hand, the large version of \textsc{SapBERT} introduces a pretraining metric learning framework grounded in self-supervised masked language modeling. It captures fine-grained semantic relationships by clustering synonyms under the same concept, learning to align biomedical entities directly from raw text without complex hybrid tuning components \citep{xu-etal-2020-generate,ji2020bert,sung2020biomedical}.

\vspace{0.53em}

The large scale of the alignment task imposes a significant computational cost, laying the groundwork for a bottleneck. For this reason, we propose an interoperable cutting-edge optimization process focused on quantization, as introduced in \Cref{fig:front}. Fundamentally, the performance of alignment techniques is closely linked with time requirements and computational resource limitations, making efficiency-critical optimizations essential. Accordingly, \textsc{Microsoft Olive} is leveraged to intelligently search for optimizations among different Execution Providers (EPs) using the \textsc{ONNX Runtime} backend. Sequentially, an accuracy-preserving quantization is then applied using \textsc{Intel Neural Compressor} and \textsc{IPEX}, along with \textsc{SmoothQuant} \citep{xiao2024smoothquantaccurateefficientposttraining}, shifting complexity from activations to weights. This involves engineering the scaling factor matrix $S$ and the smoothing factor $\alpha$ to mathematically resolve both the dequantization complexity and the inherent incompatibility with accelerated hardware kernels, which cannot tolerate lower-throughput operations. 

\vspace{0.53em}

To further assess the optimization impact, we systematically conduct calibration procedures using diverse biomedical datasets, focusing on terminology alignment. We then quantify efficiency on the two DEFT 2020 benchmark tasks \citep{cardon-etal-2020-presentation}, which closely match our research objective and allow a rigorous analysis of trade-off metrics.

\clearpage
\section{Related Work}\label{sec:related}
\textbf{Biomedical Ontology Alignment.} Since knowledge source builders concerned with developing health systems for various model organisms joined to create the Gene Ontology Consortium in 1998, the need for biomedical ontology alignment applications \citep{inbook99} has grown significantly, aiming to determine correspondences between concepts across different ontologies \citep{0018324}. Scalable logic-based ontology matching systems, including \textsc{LogMap} \citep{10.1007/978-3-642-25073-6_18} and \textsc{AgreementMakerLight} (AML) \citep{10.1007/978-3-642-41030-7_38}, treat alignment as a sequential process, starting with lexical matching, followed by mapping extension and correction. Yet these systems primarily consider surface-level text forms, ignoring word semantics.

Recent machine learning approaches, such as \textsc{DeepAlignment} \citep{kolyvakis-etal-2018-deepalignment} and \textsc{OntoEmma} \citep{wang-etal-2018-ontology}, map words into vector spaces using embeddings, where semantically closer words have smaller similarity distances. However, non-contextual embeddings limit their ability to disambiguate meaning. Fine-tuned \textsc{BERT} models \citep{He2021BiomedicalOA} and Siamese Neural Networks (\textsc{SiamNN}) \citep{10.1007/978-3-030-77385-4_23} show improved performance, but challenges remain due to limited annotated data and the large entity space.

To address these challenges, we adopt ontology alignment systems based on state-of-the-art supervised learning schemes, utilizing domain-specific knowledge from UMLS. 
Our approach combines \textsc{KRISSBERT} \citep{zhang2022knowledgerichselfsupervisionbiomedicalentity}, which effectively resolves variations and ambiguities among millions of entities through self-supervision, and the large \textsc{SapBERT} variant \citep{liu2021selfalignmentpretrainingbiomedicalentity}, which employs an extensive metric learning framework to self-align synonymous biomedical entities, linking synonyms into a unified semantic notion.
Unlike pragmatic pretrained models, notably \textsc{BioBERT} \citep{lee2020biobert}, \textsc{PubMedBERT} \citep{10.1145/3458754}, and \textsc{Bioformer} \citep{fang2023bioformerefficienttransformerlanguage}, which still require labeled data such as gold mention occurrences, constrained by annotation scarcity across expansive biomedical domains, and struggle to produce well-differentiated embedding spaces, our approach captures contextual meaning more efficiently. It coherently retrieves all UMLS entities sharing surface forms and supports the generation of distinct representations for semantically different biomedical concepts.

\stitle{Model Optimizations.} Techniques for accelerating and compressing deep learning models have garnered significant attention due to their ability to reduce parameters, computations, and energy-intensive memory access. Optimization methods in neural networks date back to the late 1980s \citep{NIPS1989_6c9882bb,10.1162/neco.1992.4.4.473}, with quantization (approximating numerical components with low bit-width precision) \citep{Jacob_2018_CVPR,wu2020integerquantizationdeeplearning,unknown}, pruning (removing less important connections to create sparse networks) \citep{NIPS1992_303ed4c6,conf/iclr/FrankleC19}, and knowledge distillation (teacher-student neural model paradigm) \citep{hinton2015distillingknowledgeneuralnetwork,Xu2017TrainingSA} becoming widely adopted. These techniques allow smaller models to operate efficiently within energy-saving on-chip memory, reducing reliance on high-latency off-chip DRAM.
Recent advances highlight the importance of combining optimization strategies for greater efficiency \citep{inbook75,b832041edd824d71a8f43fd0017ba097}. Quantization, achieving significant compression with minimal accuracy loss \citep{carreiraperpiñán2017modelcompressionconstrainedoptimization}, is often paired with pruning \citep{yu2020jointpruningquantization,Qu_2020_CVPR}, automatic mixed precision \citep{article88,rakka2022mixedprecisionneuralnetworkssurvey}, and performance tuning \citep{article56} in sequential pipelines. Extensively applied in transformers \citep{articleQ,kim2021ibertintegeronlybertquantization,schaefer2023augmentinghessiansinterlayerdependencies}, quantization benefits from techniques such as weight equalization \citep{nagel2019datafreequantizationweightequalization} and channel splitting \citep{zhao2019improvingneuralnetworkquantization}, which address weight outliers but fall short in handling activation outliers, a persistent bottleneck.
In response, our novel proposed quantization approach efficiently mitigates activation outliers by shifting the complexity to weight quantization \citep{xiao2024smoothquantaccurateefficientposttraining}.

\vspace{0.25em}

\stitle{End-to-End Hardware-aware Optimizations.} Initially, researchers focused on software optimizations before addressing hardware efficiency \citep{Han2015DeepCC,10.5555/2969442.2969588}. However, this static approach fails to exploit the dynamic potential of combining compression techniques to improve performance \citep{guo2016dynamicnetworksurgeryefficient,inproceedings44}. By optimizing memory transfers and leveraging parallelism, compressed models significantly reduce both hardware costs and resource demands \citep{inproceedings34,10226296,10380502}.
To this end, we leverage \textsc{Microsoft Olive}, with its dedicated ecosystem, to algorithmically engineer the optimization process.
\section{Methodology}\label{sec:method}
{In line with our study objective, which focuses on aligning biomedical ontologies using cosine similarity measures, we align the concatenation of two fields, \emph{Biomedical Term} and \emph{Public Explanation}, from the layman biomedical vocabulary with all the French entries in the \emph{String (ST)} field of the \texttt{MRCONSO.RRF} raw file from the AB2024 UMLS Metathesaurus release.
To accomplish this, we devised a sequential algorithmic search process designed to optimize model performance across multiple EPs.
It integrates network compression, parallel processing, and memory transfer optimization through \textsc{Microsoft Olive}, in cooperation with the \textsc{ONNX Runtime} backend, thus enabling efficient and scalable execution. 
Furthermore, within this framework, we employ \textsc{Intel Neural Compressor} and \textsc{IPEX}, incorporating the logic of \textsc{SmoothQuant}, to design a search-optimized, on-the-fly quantization strategy (W8A8). This approach uniformly shifts the burden from activation outliers to weights, thereby enhancing compatibility with specific hardware-accelerated kernels.

By adopting this strategy, memory usage is significantly reduced and inference speed improved, both critical factors for effective alignment. This synergy, essential to the performance of biomedical ontology systems, depends on these optimizations to ensure dynamic scalability.}

\stitle{Formal Definition.} 
An ontology is typically defined as an explicit specification of a conceptualization. It often uses representational vocabularies to describe a domain of interest, with the main components being entities\footnote{Entities include classes, instances, properties, relationships, data types, annotations, and cardinality constraints.} and axioms. Ontology alignment involves matching cross-ontology entities with equivalence, subsumption, or related relationships. Alongside this, the current study focuses on equivalence alignment between classes.\footnote{A class of an ontology typically contains a list of labels (via annotation properties such as \emph{rdfs:label}) that serve as alternative class names, descriptions, synonyms, or aliases.}

The ontology alignment system inputs a pair of ontologies, $O$ and $O'$, with class sets $C$ and $C'$. It generates, using cosine similarity, a set of scored mappings in the form $(c \in C, c' \in C', P(c \equiv c'))$, where $P(c \equiv c') \in [0,1]$ is the probability score (\emph{mapping value}) of equivalence between $c$ and $c'$. Final mappings are selected based on the highest scores, leveraging supervised SOTA learning schemes with feature engineering. When one model produces more accurate alignments, these are used to correct those of the other, with manual verification by human annotators for reliability.

In the present architecture, the input sequence includes a special token \texttt{[CLS]}, the tokens of two sentences $A$ and $B$, and the special token \texttt{[SEP]} separating them. Each token embedding encodes its content, position, and sentence information. 
In $\mathcal{L}$ successive layers of the architecture, the multi-head self-attention block computes contextualized representations for each token. The output of layer $l$ is the embedding sequence derived from the input, as defined in Eq.~\eqref{eq:embedding_output}:
\vspace{0.22em}
\begin{equation}
\label{eq:embedding_output}
\begin{split}
    f_{bert}(\mathbf{x}, l) = &\; (\mathbf{v}_{CLS}^{(l)}, \mathbf{v}_1^{(l)}, \ldots, \mathbf{v}_N^{(l)}, \\ 
    &\;\;\;\;\mathbf{v}_{SEP}^{(l)}, \mathbf{v}_1'^{(l)}, \ldots, \mathbf{v}_{N'}'^{(l)}) \\ 
    &\;\;\;\;\in \mathbb{R}^{(N+N'+2) \times d}
\end{split}
\end{equation}
where $\mathbf{x}$ is the input sequence, $\mathbf{v}_i^{(l)}$ and $\mathbf{v}_j'^{(l)}$ are $d$-dimensional vectors of the respective tokens. The final layer ($l = \mathcal{L}$) outputs the resulting token embeddings. Unlike non-contextual embeddings such as Word2Vec \citep{DBLP:journals/corr/abs-1301-3781}, which assign one embedding per token, this configuration distinguishes occurrences of the same token in different contexts. This is critical in expanding biomedical domains where traditional embeddings are biased towards frequent meanings in training corpora.
For instance, depending on the context, "MS" can refer to \emph{Multiple Sclerosis}, a chronic neurological disease, or \emph{Mass Spectrometry}, an analytical method for measuring ion mass-to-charge ratios.

Concordantly, given input ontologies $O$ and $O'$ with class sets $C$ and $C'$, a naive algorithm computes alignments by looking up $c' = \arg\max_{c' \in C'} P(c \equiv c')$ for each $c \in C$, leading to $O(n^2)$ time complexity. This is parametrically enhanced via \textsc{Microsoft Olive}, which employs an optimal search approach that calibrates a \texttt{joint}\footnote{Search spaces of all passes are combined and jointly evaluated to find optimal parameters, using Optuna's TPESampler.} execution order, backed by the \texttt{TPE} (Tree-structured Parzen Estimator) algorithm.

\begin{figure}[H]
\vspace{0.193em}
\centering
    \includegraphics[width=\linewidth]{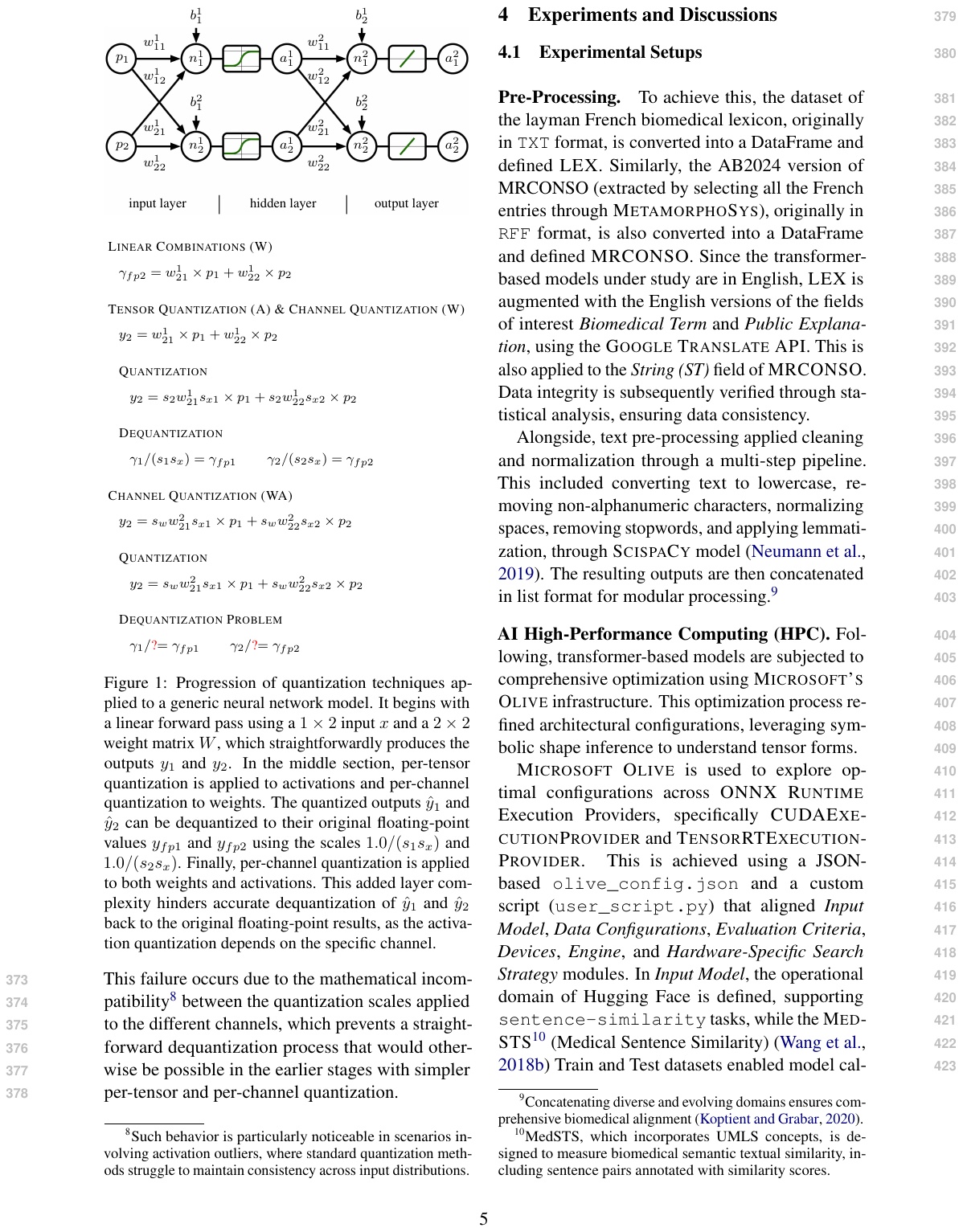}
    \caption{
Progression of quantization techniques applied to a generic neural network model. It begins with a linear forward pass using a \(1 \times 2\) input \( x \) and a \(2 \times 2\) weight matrix \( W \), which produces the outputs \( y_1 \) and \( y_2 \) in a straightforward floating-point manner.
In the middle section, per-tensor quantization is performed on activation outputs, and per-channel quantization on weights. The quantized outputs \( \hat{y}_1 \) and \( \hat{y}_2 \) can be dequantized to their original floating-point values \( y_{fp1} \) and \( y_{fp2} \) using the channel-specific scales \( 1.0 / (s_1 s_x) \) and \( 1.0 / (s_2 s_x) \), respectively.
Finally, both weights and activations undergo per-channel quantization. This additional layer of complexity hinders accurate dequantization of \( \hat{y}_1 \) and \( \hat{y}_2 \) back to their original floating-point results, as the activation quantization depends on the specific channel.}
    \label{fig:schema}
    \vspace{-1em}
\end{figure}

Our search-optimized quantization pipeline (W8A8) further improves efficiency by shifting computational complexity from activations to weights, ensuring seamless integration with hardware-accelerated compute units and resolving\footnote{This outcome involves \texttt{Mul} operations without folding, optimized in \textsc{IPEX} through system-level automatic fusion.} dequantization issues, conforming to \Cref{fig:schema}.

\begingroup
\noindent
The present failure occurs due to the mathematical incompatibility\footnote{Such behavior is particularly noticeable in scenarios involving activation outliers, where standard quantization methods struggle to maintain consistency across input distributions.} between the quantization scales applied to the different channels, which prevents a straightforward dequantization process that would otherwise be possible in the earlier stages with simpler per-tensor and per-channel quantization.
\endgroup

\begingroup
\setlength{\parskip}{0pt}

\subsection{Mathematical Model} \label{sec:math}

Following optimization, the dynamically quantized model, together with the tokenizer  
\(\mathcal{T} : \mathcal{D} \to \mathbb{R}^{B \times L \times D}\), is loaded, where \(\mathcal{D}\) denotes the set of raw text inputs, \(B\) the batch size, \(L\) the sequence length, and \(D\) the embedding dimension.

In turn, a batch-encoding function is introduced to process the lists of interest. It initializes data structures for collecting text-batch embeddings and temporarily stores intermediate results to streamline alignment mechanisms. This ensures that subsequent computations are performed efficiently, improving throughput and avoiding memory bottlenecks during batch processing.

The set of texts \(\mathbf{T} = \{T_1, T_2, \dots, T_N\}\), with \(N = |\mathbf{T}|\), is divided into batches of size \(B = 10\), denoted \(\mathbf{B}_k\) for \(k = 1, \dots, K\), where \(K = \lceil \frac{N}{B} \rceil\), as formulated in Eq.~\eqref{eq:batch_union}:
\begin{equation}
\label{eq:batch_union}
    \mathbf{T} = \bigcup_{k=1}^{K} \mathbf{B}_k
\end{equation}
Each batch \(\mathbf{B}_k\) is defined as in Eq.~\eqref{eq:batch_definition}:
\begin{equation}
\label{eq:batch_definition}
\setlength{\abovedisplayskip}{12pt}
\setlength{\belowdisplayskip}{12pt}
\begin{split}
    \mathbf{B}_k = &\; \{ T_{(k-1)B + 1}, T_{(k-1)B + 2}, \dots, \\ 
    &\;\;\;\;\;\;\;\;\;\;\;\;\;\;\;\;\;\;\;\;\;\;\;\;\;\; T_{\min(kB, N)} \}
\end{split}
\end{equation}

Accordingly, the tokenizer \(\mathcal{T}\) maps the textual input in each batch \(\mathbf{B}_k\) to its numerical tensor representation \(\mathbf{X}_k\), as established in Eq.~\eqref{eq:tokenization}:
\begin{equation}
\label{eq:tokenization}
    \mathbf{X}_k = \mathcal{T}(\mathbf{B}_k)
\end{equation}
where the tokenized data \(\mathbf{X}_k \in \mathbb{R}^{B \times L \times D}\) represents each batch. Thus, padding and truncation ensure uniform sequence lengths, with \(L = 512\) set via the \texttt{max\_length} parameter. The resulting outputs are converted into PyTorch tensors, enabling consistent formatting across batches. This standardization reinforces compatibility and integration with ONNX-based pipelines, after which the tensors are cast to NumPy arrays for seamless transfer within the processing infrastructure.

\textsc{ONNX Runtime} is then activated by initiating a session that processes the dynamically quantized model  
\(\mathcal{M} : \mathbb{R}^{B \times L \times D} \to \mathbb{R}^{B \times L \times H}\), producing the embeddings \(\mathbf{H}_k\), given by Eq.~\eqref{eq:model_forward}:
\begin{equation}
\label{eq:model_forward}
    \mathbf{H}_k = \mathcal{M}(\mathbf{X}_k)
\end{equation}
where $\mathbf{H}_k = [\mathbf{h}_{kij}] \in \mathbb{R}^{B \times L \times H}$, with $\mathbf{h}_{kij} \in \mathbb{R}^{H}$ denoting the hidden-state vector corresponding to the $j$-th token of the $i$-th input in batch $k$, and $H$ denoting the model’s output hidden dimension.

Embeddings are then converted into PyTorch tensors and averaged across the sequence length to produce fixed-size, batch-level representations, in accordance with Eq.~\eqref{eq:mean_pooling}:
\begin{equation}
\label{eq:mean_pooling}
    \mathbf{e}_{ki} = \frac{1}{L} \sum_{j=1}^{L} \mathbf{h}_{kij}
\end{equation}
This yields \(\mathbf{E}_k \in \mathbb{R}^{B \times H}\), where each row \(\mathbf{e}_{ki}\) corresponds to the mean-pooled embedding of a single input in batch \(k\).  
The final dataset-level embedding matrix \(\mathbf{E} \in \mathbb{R}^{N \times H}\) is then obtained by stacking all individual embedding vectors \(\mathbf{e}_i^\top \in \mathbb{R}^{1 \times H}\) (for \(i = 1, \dots, N\)), which are grouped into the batch-level matrices \(\mathbf{E}_k\) (for \(k = 1, \dots, K\)), as in Eq.~\eqref{eq:concat_embeddings}:
\begin{equation}
\label{eq:concat_embeddings}
    \mathbf{E} =
    \begin{bmatrix}
        \mathbf{e}_1^\top \\
        \mathbf{e}_2^\top \\
        \vdots \\
        \mathbf{e}_N^\top
    \end{bmatrix}
    =
    \begin{bmatrix}
        \mathbf{E}_1 \\
        \vdots \\
        \mathbf{E}_K
    \end{bmatrix}
\end{equation}
Using this function, two sets of texts are encoded, as specified in Eq.~\eqref{eq:encode_embeddings}, producing the embedding tensors \(\mathbf{E}_{\mathit{L}}\) and \(\mathbf{E}_{\mathit{M}}\), where \(\mathbf{L} = \{T_{L_1}, \dots, T_{L_{N_L}}\}\) and \(\mathbf{M} = \{T_{M_1}, \dots, T_{M_{N_M}}\}\) are the input collections from LEX and MRCONSO, respectively:
\begin{equation}
\label{eq:encode_embeddings}
\begin{aligned}
    \mathbf{E}_{\mathit{L}} &= \text{EncodeBatch}(\mathbf{L}) \in \mathbb{R}^{N_{\mathit{L}} \times H} \\
    \mathbf{E}_{\mathit{M}} &= \text{EncodeBatch}(\mathbf{M}) \in \mathbb{R}^{N_{\mathit{M}} \times H}
\end{aligned}
\end{equation}

Cosine similarity is then computed to quantify pairwise semantic similarity between embeddings. For two vectors \(\mathbf{a}\) and \(\mathbf{b}\), it is defined as in Eq.~\eqref{eq:cosine_similarity}:
\begin{equation}
\label{eq:cosine_similarity}
    \text{cosine\_similarity}(\mathbf{a}, \mathbf{b}) = \frac{\mathbf{a}^\top \mathbf{b}}{\|\mathbf{a}\|_2 \|\mathbf{b}\|_2}
\end{equation}
The resulting matrix \(\mathbf{S} \in \mathbb{R}^{N_L \times N_M}\), where each element \((i,j)\) represents the similarity between the \(i\)-th embedding vector \(\mathbf{E}_{Li} \in \mathbb{R}^H\) in LEX and the \(j\)-th embedding vector \(\mathbf{E}_{Mj} \in \mathbb{R}^H\) in MRCONSO, is given in Eq.~\eqref{eq:similarity_matrix}:
\begin{equation}
\label{eq:similarity_matrix}
\setlength{\abovedisplayskip}{12pt}
\setlength{\belowdisplayskip}{12pt}
    \begin{gathered}
        \mathbf{S}_{ij} = \text{cosine\_similarity}(\mathbf{E}_{Li}, \mathbf{E}_{Mj}) \\ 
        = \frac{\mathbf{E}_{Li}^\top \mathbf{E}_{Mj}}{\|\mathbf{E}_{Li}\|_2 \|\mathbf{E}_{Mj}\|_2}
    \end{gathered}
\end{equation}

Finally, each term \(T_{L_i}\) in LEX is aligned to its closest semantic counterpart in MRCONSO by selecting the index \(j_i^*\) that maximizes the cosine similarity, as determined in Eq.~\eqref{eq:argmax_alignment}:
\begin{equation}
\label{eq:argmax_alignment}
    j_i^* = \arg\max_j \mathbf{S}_{ij}
\end{equation}

\section{Experiments and Discussions}\label{sec:experiment}
\subsection{Experimental Setups}

\vspace{-0.65em}

\stitle{Preprocessing.} \label{sec:preprocessing}
To achieve this, the dataset of the French layman biomedical lexicon, originally in \texttt{TXT} format, is converted into a DataFrame and defined as LEX. Similarly, the AB2024 version of MRCONSO (extracted by selecting all French entries via \textsc{MetamorphoSys}), originally in \texttt{RRF} format, is also converted into a DataFrame and referred to as MRCONSO. Since the transformer-based models under study are in English, LEX is augmented with the English translations of the fields of interest \emph{Biomedical Term} and \emph{Public Explanation}, using the \textsc{Google Translate API}. The same translation is applied to the \emph{String (ST)} field of MRCONSO. Data integrity is then verified through statistical analysis, assessing distributional properties, missing values, and outliers.

Subsequently, text preprocessing is performed via a multi-step pipeline of cleaning and normalization. This includes converting text to lowercase, removing non-alphanumeric characters, normalizing spaces, removing stopwords, and applying lemmatization through the \textsc{ScispaCy} model \citep{neumann-etal-2019-scispacy}. The resulting outputs are concatenated into a list format for modular processing.\footnote{The concatenation of evolving domains ensures comprehensive biomedical alignment \citep{koptient:hal-03095275}.}

\stitle{AI High-Performance Computing (HPC).}  
The transformer-based models undergo comprehensive optimization via the infrastructure of \textsc{Microsoft Olive}. This optimization process refines architectural configurations by leveraging symbolic shape inference to understand tensor shapes.

\textsc{Microsoft Olive} is used to explore optimal configurations across \textsc{ONNX Runtime} Execution Providers, specifically \textsc{CUDAExecutionProvider} and \textsc{TensorRTExecutionProvider}. This is achieved using a JSON-based configuration file (\texttt{olive\_config.json}) and a custom script (\texttt{user\_script.py}) that configures the \emph{Input Model}, \emph{Data Configurations}, \emph{Evaluation Criteria}, \emph{Devices}, \emph{Engine}, and \emph{Search Strategy} modules. In \emph{Input Model}, the operational domain of Hugging Face is defined, supporting the \texttt{sentence-similarity} task, while the MedSTS\footnote{MedSTS, which incorporates UMLS concepts, is designed to measure biomedical semantic textual similarity, including sentence pairs annotated with similarity scores.} (Medical Sentence Similarity) \citep{journals/corr/abs-1808-09397} Train and Test datasets serve as resources for model calibration through the \emph{Data Configurations} module. \emph{Evaluation Criteria} include accuracy, precision, recall, F1-score, and latency (average, maximum, minimum). The cache directories manage intermediate results, streamlining reproducibility and scalability. Optimization goals are defined algorithmically and adhered to strict parametric thresholds: a maximum performance degradation of \(0.01\%\) and a minimum latency improvement of \(20\%\). In the \emph{Device} module, \texttt{local\_system} is designated as the GPU-supported system. \emph{Engine and Search Strategy} employ the \texttt{joint} execution order with the \texttt{TPE} algorithm, for profiling within the search space.

\stitle{ONNX Runtime Passes.}  
Optimization begins with \emph{OnnxConversion}, which converts PyTorch models to \textsc{ONNX} format (\texttt{opset: 14}) for hardware-agnostic execution. Subsequently, \emph{OrtTransformersOptimization} module streamlines computational graphs by combining adjacent layers and pruning redundant nodes. \emph{OrtMixedPrecision} enhances throughput and reduces memory usage by performing FP16\footnote{Float16 precision is enabled for \textsc{CUDAExecutionProvider} but disabled for \textsc{TensorRTExecutionProvider}, balancing compatibility and computational gains.} arithmetic where applicable. Lastly, \emph{OrtPerfTuning} profiles latency and throughput, performing runtime tuning\footnote{The proposed runtime tuning enhances model calibration and inference through dynamic architectural optimization.} in model configurations. The sequential application of these optimization steps enables modular result storage, allowing model assessment via Pareto frontier analysis.

\stitle{Search-Optimized Quantization.}  
The INT8 (W8A8) quantization logic is implemented using \textsc{SmoothQuant} \citep{xiao2024smoothquantaccurateefficientposttraining}, coordinating \textsc{Intel Neural Compressor} and \textsc{IPEX} (Intel Extension for PyTorch), together with \textsc{Microsoft Olive} and the \textsc{ONNX Runtime} backend. The \emph{QOperator} format includes \emph{QLinearMatMul}, \emph{MatMulInteger}, \emph{QLinearAdd}, and \emph{QLinearRelu} operators, configured via custom JSON settings, in order to manage the transversal redistribution of quantization complexity through a smoothing factor $\alpha=0.5$, validated as optimal for the models from Microsoft Research and Cambridge LTL. The use of NGC containers streamlines the integration of the previous configuration script (\texttt{user\_script.py}) and calibration datasets, to ensure scalable model deployment on accelerated hardware, while retaining optimization objectives.

\subsection{Main Results and Analysis} 

\vspace{-0.65em}

\stitle{DEFT 2020 Evaluation Campaign.} \label{sec:deft_2020}
Since, in our case study, there is no test dataset for inference matched with a training dataset for calibration, the MedSTS resources are used for this purpose, and inference is applied directly to this end as part of our approach. In addition, to quantify the efficiency of our optimization processes by means of performance, latency, and consumption metrics, we use the datasets from the two tasks of the DEFT 2020 Evaluation Campaign \citep{cardon-etal-2020-presentation}, as they are broadly representative of our core objective of biomedical ontology alignment.\footnote{In the Train module, the pretrained models are calibrated by framing optimal model optimizations aligned with the highest hardware performance capabilities, whereas in the Test module, the evaluation metrics are established.}

In Task 1, which aims to identify the degree of semantic similarity between pairs of sentences, the \texttt{input\_cols} parameter is set to \texttt{[sentence1, sentence2]}, corresponding to the \emph{source} and \emph{target} fields, respectively. These are formatted as token sequences, and the \texttt{label\_cols} parameter is set to \texttt{[label]} for the \emph{mark} field, representing human-assigned scores from 0 to 5 indicating pairwise sentence-level semantic correspondence.

The same functional topology is transversally adapted for Task 2, concerning the identification of parallel sentences.\footnote{The parallelism of the sentences is related to the simple-complex relationship, ergo one of the simple sentences (\emph{target}) is always derived from the complex sentence (\emph{source}).} In turn, the data from the latter are internally linked with the corresponding identifier present in the \emph{num} field. This linkage linearly maps the inferential string yielding the highest cosine similarity score for each virtually tripartitioned segment, created based on the associated \emph{id} of each data line. Thus, the correspondence with the identifier in \texttt{[label]}, representing the \emph{target} field, is ensured.
The adoption of virtual compartment systems with three distinct conditions is introduced because the second task aims to identify, among three \emph{target} sentences, the one that best corresponds to the \emph{source} in terms of sentence-level parallelism.

\stitle{Configurational Decorators.} These configuration architectures are diligently designed using logging wrappers (decorators) to log the methodically engineered processing pipeline, and to generate the dataloader through \textsc{HuggingfaceDataContainer}. In practical application, this component enables robust evaluation metrics testing, thereby presenting a wide range of potential options.

\stitle{Task 1.} \label{sec:task_1} The first task, focused on continuous semantic evaluation (Semantic Similarity Evaluation), presented complications in converting the models' inference outputs from cosine similarity percentages to the compliant evaluation format. Specifically, it has been found that, particularly for \textsc{KRISSBERT} \citep{zhang2022knowledgerichselfsupervisionbiomedicalentity}, the percentage scores of cosine semantic similarity are extremely high compared to the norm. This is presumably due to an improperly calibrated cross-entropy loss in the training of the cross-attention encoder, as cursorily reported in Microsoft Research’s study, which results in the re-ranking score being maximized even for partial or incorrect entities. The model's inferences, while excelling in Named Entity Linking (NEL), lead to problems in cosine similarity score attribution. It is also advisable to review the linear layer applied to the encoding of the first \texttt{[CLS]} token to calculate the re-ranking score, as it has been proven that the score is very high even for nonsensical sentence pairs, potentially indicating poor discrimination. To address this, a feature scaling function using \texttt{MinMaxScaler} is manually added in the \texttt{post\_process\_data} module of \textsc{HuggingfaceDataContainer}, converging into a corrective fine-tuning (see \Cref{tab:improvement}). This enabled the use of the official EDRM evaluation metric \citep{cardon-etal-2020-presentation}, which measures the average relative distance to the solution as a micro-average. For each similarity value, the reference data \( r_i \) corresponds to the maximum possible distance between the system's predicted response and the data \( d_{\text{max}}(h_i, r_i) \), formally defined in Eq.~\eqref{eq:edrm}:
\vspace{0.01em}
\begin{equation}
\label{eq:edrm}
\text{EDRM} = \frac{1}{n} \sum_{i=1}^{n} \left(1 - \frac{d(h_i, r_i)}{\text{dmax}(h_i, r_i)} \right)
\end{equation}

Our technique surpassed the previous FP32 state-of-the-art achieved by UASZ (Université Assane Seck de Ziguinchor) \citep{drame-etal-2020-approche}, as presented in \Cref{tab:task1SOTA}, and more statistically in \Cref{fig:regression}.

\begin{table}[H]
\setlength{\tabcolsep}{3pt}
\centering
\scriptsize
\vspace{0em}
\begin{tabular}{lccccccccccccccccccc}
\toprule
& \multicolumn{3}{c}{\small Task $@1$} \\
\cmidrule[1.5pt]{2-4}
\multirow{2}{*}{\raisebox{7\height}{\small Method}} & \multicolumn{1}{c}{EDRM} & \multicolumn{1}{c}{Spearman correlation} & \multicolumn{1}{c}{$p$-value} \\
\cmidrule[1.0pt]{1-13}
\textsc{KRISSBERT INT8} & \textbf{0.8604} & 0.8253 & 2.0724e-97 \\
\textsc{SapBERT-Large INT8} & 0.8593 & \textbf{0.8289} & \textbf{2.5965e-99} \\
\cmidrule[1.0pt]{1-18}
\textsc{UASZ} \citep{drame-etal-2020-approche}, 1 & 0.7947 & 0.7528 & 4.3371e-76 \\
\textsc{UASZ} \citep{drame-etal-2020-approche}, 2 & \textbf{0.8217} & 0.7691 & 2.3769e-81 \\
\textsc{UASZ} \citep{drame-etal-2020-approche}, 3 & 0.7755 & \textbf{0.7769} & \textbf{5.5766e-84} \\
\bottomrule
\end{tabular}
\vspace{0em}
\caption{Comparison of the study models, optimized to INT8 (W8A8) by \textsc{Microsoft Olive}, against the UASZ state-of-the-art \citep{drame-etal-2020-approche}. The metrics include EDRM, Spearman correlation, and $p$-values.}
\label{tab:task1SOTA}
\vspace{-0.8em}
\end{table}

\begin{figure}[H]
\vspace{0.5em}
\centering
    \includegraphics[width=\linewidth]{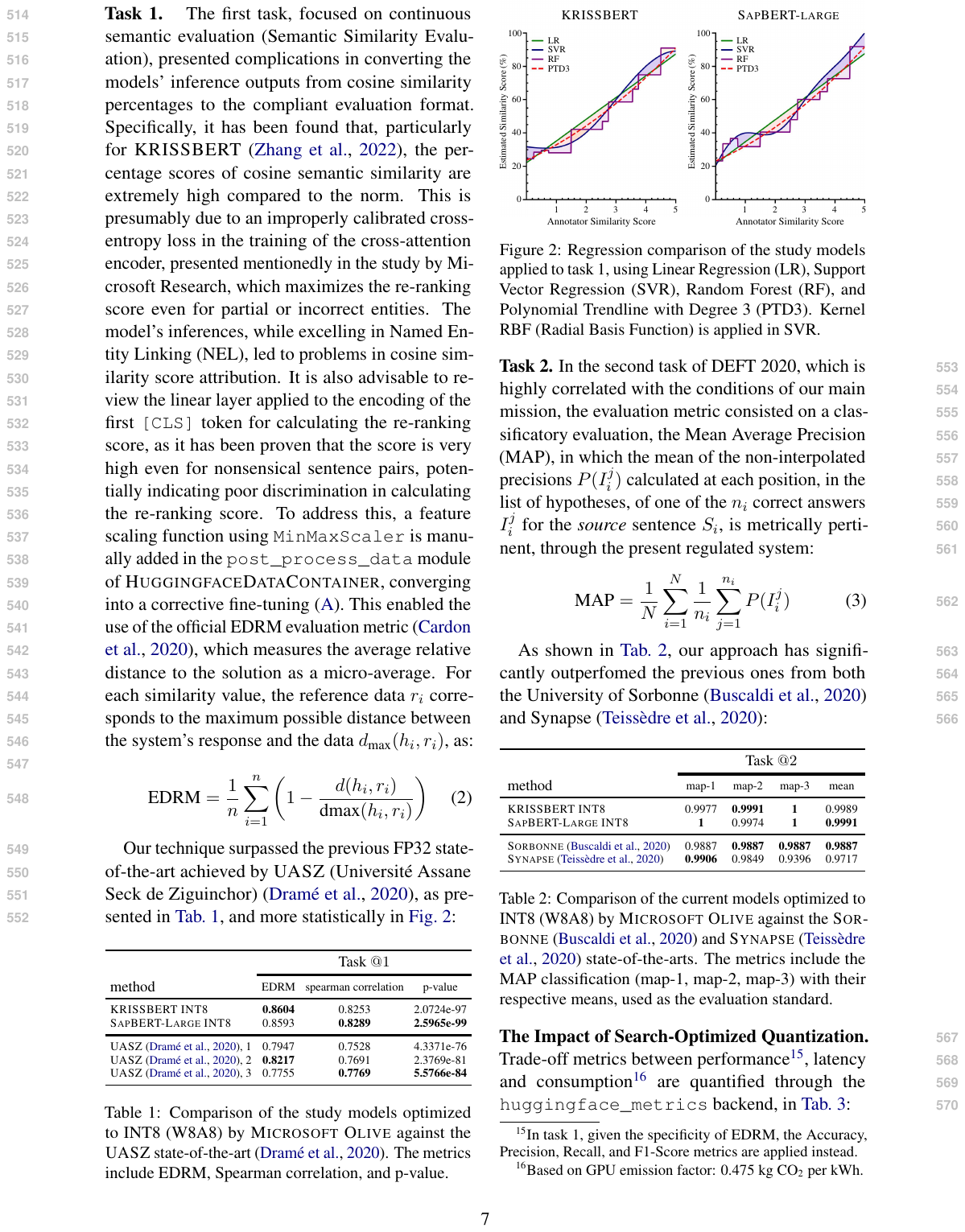}
    \vspace{-1.4em}
    \caption{
Regression comparison of the study models applied to Task 1, using Linear Regression (LR), Support Vector Regression (SVR), Random Forest (RF), and a Polynomial Trendline with Degree 3 (PTD3). The Radial Basis Function (RBF) is applied in the SVR.} 
    \label{fig:regression}
    \vspace{-1.4em}
\end{figure}

\stitle{Task 2.} In the second task of DEFT 2020, which closely aligns with the conditions of our main mission, the evaluation metric consists of a classification-based assessment: the Mean Average Precision (MAP), formulated in Eq.~\eqref{eq:map}, is computed as the mean of the non-interpolated precisions $P(I_i^j)$ at each position in the ranked list of hypotheses, for each of the $n_i$ correct answers $I_i^j$ associated with a given \emph{source} sentence $S_i$:
\vspace{-0.31em}

\begin{equation}
\text{MAP} = \frac{1}{N} \sum_{i=1}^{N} \frac{1}{n_i} \sum_{j=1}^{n_i} P(I_i^j)
\label{eq:map}
\end{equation}

As detailed in \Cref{tab:task2SOTA}, our approach has significantly outperformed the previous ones from both the University of Sorbonne \citep{buscaldi-etal-2020-calcul} and Synapse \citep{teissedre-etal-2020-similarite}.

\begin{table}[H]
\setlength{\tabcolsep}{4pt}
\centering
\scriptsize
\vspace{0em}
\begin{tabular}{lccccc}
\toprule
& \multicolumn{4}{c}{\small Task $@2$} \\
\cmidrule[1.5pt]{2-5}
\multirow{2}{*}{\raisebox{7\height}{\small Method}} & \multicolumn{1}{c}{MAP-1} & \multicolumn{1}{c}{MAP-2} & \multicolumn{1}{c}{MAP-3} & \multicolumn{1}{c}{Mean} \\
\cmidrule[1.0pt]{1-6}
\textsc{KRISSBERT INT8} & 0.9977 & \textbf{0.9991} & \textbf{1} & 0.9989 \\
\textsc{SapBERT-Large INT8} & \textbf{1} & 0.9974 & \textbf{1} & \textbf{0.9991} \\
\cmidrule[1.0pt]{1-6}
\textsc{Sorbonne} \citep{buscaldi-etal-2020-calcul} & 0.9887 & \textbf{0.9887} & \textbf{0.9887} & \textbf{0.9887} \\
\textsc{Synapse} \citep{teissedre-etal-2020-similarite} & \textbf{0.9906} & 0.9849 & 0.9396 & 0.9717 \\
\bottomrule
\end{tabular}
\vspace{0em}
\caption{Comparison of the study models, optimized to INT8 (W8A8) by \textsc{Microsoft Olive}, against the state-of-the-art benchmarks from Sorbonne \citep{buscaldi-etal-2020-calcul} and Synapse \citep{teissedre-etal-2020-similarite}. The metrics include MAP classification scores (MAP-1, MAP-2, MAP-3) with their respective mean values.}
\label{tab:task2SOTA}
\vspace{-2.1em}
\end{table}

\begin{table*}[t]
\setlength{\tabcolsep}{2.75pt}
\centering
\scriptsize
\begin{tabular}{lccccccccccccccccccc}
\toprule
& \multicolumn{4}{c}{\small Performance} & & \multicolumn{3}{c}{\small Latency} & & \multicolumn{3}{c}{\small Consumption}\\
\cmidrule[1.5pt]{2-5} \cmidrule[1.5pt]{7-9} \cmidrule[1.5pt]{11-13}
\rowcolor[gray]{0.9} \multirow{2}{*}{\raisebox{7\height}{\small Task $@1$}} & \multicolumn{1}{c}{Accuracy} & \multicolumn{1}{c}{Precision} & \multicolumn{1}{c}{Recall} & \multicolumn{1}{c}{F1-score} & & \multicolumn{1}{c}{Latency-avg} & \multicolumn{1}{c}{Latency-max} & \multicolumn{1}{c}{Latency-min} & & \multicolumn{1}{c}{Size} & \multicolumn{1}{c}{GPU energy} & \multicolumn{1}{c}{CO2} \\
\cmidrule[1.0pt]{1-20}
\textsc{KRISSBERT} \citep{zhang2022knowledgerichselfsupervisionbiomedicalentity} & 0.8886 & 0.9047 & 0.8920 & 0.8983 & & 19.9143 & 20.2043 & 19.6533 & & 438 & 2.2127 & 1.0510 \\
{\ \ \ \ + \textsc{Microsoft Olive}} & \CC{15}0.8886 & \CC{15}0.9047 & \CC{15}0.8920 & \CC{15}0.8983 & & \CCG{30}1.2114 & \CCG{30}1.2165 & \CCG{30}1.2051 & & \CCG{30}166.44 & \CCG{30}0.1346 & \CCG{30}0.0639 \\
\cmidrule[1.0pt]{1-18}
\textsc{SapBERT-large} \citep{liu2021selfalignmentpretrainingbiomedicalentity} & 0.8808 & 0.8851 & 0.8937 & 0.8894 & & 64.0251 & 64.3159 & 63.7649 & & 2293.76 & 7.1139 & 3.3791 \\
{\ \ \ \ + \textsc{Microsoft Olive}} & \CC{15}0.8808 & \CC{15}0.8851 & \CC{15}0.8937 & \CC{15}0.8894 & & \CCG{30}3.0494 & \CCG{30}3.0562 & \CCG{30}3.0453 & & \CCG{30}756.94 & \CCG{30}0.3388 & \CCG{30}0.1609 \\
\midrule[1.5pt] 
\rowcolor[gray]{0.9} \multirow{2}{*}{\raisebox{7\height}{\small Task $@2$}} & \multicolumn{1}{c}{MAP-1} & \multicolumn{1}{c}{MAP-2} & \multicolumn{1}{c}{MAP-3} & \multicolumn{1}{c}{Mean} & & \multicolumn{1}{c}{Latency-avg} & \multicolumn{1}{c}{Latency-max} & \multicolumn{1}{c}{Latency-min} & & \multicolumn{1}{c}{Size} & \multicolumn{1}{c}{GPU energy} & \multicolumn{1}{c}{CO2} \\
\cmidrule[1.0pt]{1-20}
\textsc{KRISSBERT} \citep{zhang2022knowledgerichselfsupervisionbiomedicalentity} & 0.9977 & 0.9991 & 1 & 0.9989 & & 55.3579 & 55.6289 & 55.1095 & & 438 & 6.1509 & 2.9217 \\
{\ \ \ \ + \textsc{Microsoft Olive}} & \CC{15}0.9977 & \CC{15}0.9991 & \CC{15}1 & \CC{15}0.9989 & & \CCG{30}3.0276 & \CCG{30}3.0351 & \CCG{30}3.0228 & & \CCG{30}171.58 & \CCG{30}0.3364 & \CCG{30}0.1598 \\
\cmidrule[1.0pt]{1-18}
\textsc{SapBERT-large} \citep{liu2021selfalignmentpretrainingbiomedicalentity} & 1 & 0.9974 & 1 & 0.9991 & & 185.5632 & 185.8308 & 185.3122 & & 2293.76 & 20.6181 & 9.7936 \\
{\ \ \ \ + \textsc{Microsoft Olive}} & \CC{15}1 & \CC{15}0.9974 & \CC{15}1 & \CC{15}0.9991 & & \CCG{30}9.7195 & \CCG{30}9.7255 & \CCG{30}9.7138 & & \CCG{30}762.13 & \CCG{30}1.0799 & \CCG{30}0.5130 \\
\bottomrule
\end{tabular}
\caption{Comparison of performance, latency, and consumption metrics for \textsc{KRISSBERT} and \textsc{SapBERT-large} models before and after optimization across the two tasks of the DEFT 2020 Evaluation Campaign. \colorbox{blue!15}{Blue} indicates maintained performance metrics in both the original and the algorithm-driven optimized models, while the transition to \colorbox{green!30}{Green} indicates improvements in both timing and resource utilization. In both cases, the optimization process yields reduced latency and energy consumption, while preserving overall performance. All results refer to inference.}
\label{tab:combined_opt_vs_deprecated}
\vspace{-1.4em}
\end{table*}

\stitle{The Impact of Search-Optimized Quantization.}
Trade-off metrics among performance\footnote{In Task 1, the specificity of the EDRM metric requires the use of accuracy, precision, recall, and F1-score.}, latency, power consumption, and estimated carbon emissions\footnote{A GPU emission factor of 0.475 kg CO\textsubscript{2}/kWh is assumed.} are quantified, as reported in \Cref{tab:combined_opt_vs_deprecated}.

For observational purposes, the effectiveness of the process is validated using the \emph{Quantization Debug} module of \textsc{ONNX Runtime}, which provides a detailed graphical representation of the redistribution of computational complexity.\footnote{The module handles activation outliers, which commonly fall within the absolute value range of 2.5 to 5, with extreme cases peaking above 7.5, thus affecting scaling factors.} For simplicity, the comparison between the activation tensors from the original computation graph and its quantized counterpart is demonstrated in \Cref{fig:dynamicQT}.
\begin{figure}[H]
\centering
    \vspace{-0.5em}
    \includegraphics[width=\linewidth]{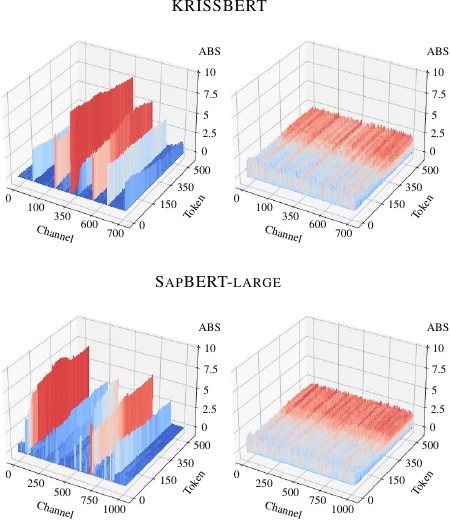}
    \vspace{-1.25em}
    \caption{
Impact of search-optimized quantization on the distribution of activations in the models under study, before and after optimization. Several channels in the original activation map display significantly high magnitudes, while the variance within a particular activation channel is consistently and notably low throughout.}
    \label{fig:dynamicQT}
    \vspace{-1.0em}
\end{figure}

\stitle{Biomedical Ontology Alignment.}
Upon completion of the vocabulary alignment, the manual verification is performed using the six-point rating scale. The results are reported in \Cref{tab:alignment-distribution}, followed by a Gaussian analysis in \Cref{fig:ontology_alignment_evaluation}, highlighting overall performance consistency across model formats.

\begin{table}[H]
\setlength{\tabcolsep}{4pt}
\centering
\scriptsize
\vspace{0em}
\begin{tabular}{lcccccc}
\toprule
\small Model & $@0$ & $@1$ & $@2$ & $@3$ & $@4$ & $@5$ \\
\cmidrule[1pt]{1-7}
\textsc{KRISSBERT INT8}       & 186  & 798  & 1,343 & 3,028 & 4,098 & 7,941 \\
\textsc{SapBERT-Large INT8}   & 205  & 687  & 1,403 & 2,928 & 4,169 & 8,002 \\
\addlinespace[0.2em]
\hdashline
\addlinespace[0.4em]
\textsc{\ \ \ \ + Complementarity}   & /  & /  & / & 897 & 5,473 & 11,024 \\
\bottomrule
\end{tabular}
\vspace{0em}
\caption{Comparison of manual rating distributions over scores $@k$ for vocabulary alignments across individual models and their complementary combination.}
\label{tab:alignment-distribution}
\vspace{-2em}
\end{table}

\vspace{1.6em}

\begin{figure}[H]
\centering
    \vspace{-1.2em}
    \includegraphics[width=\linewidth]{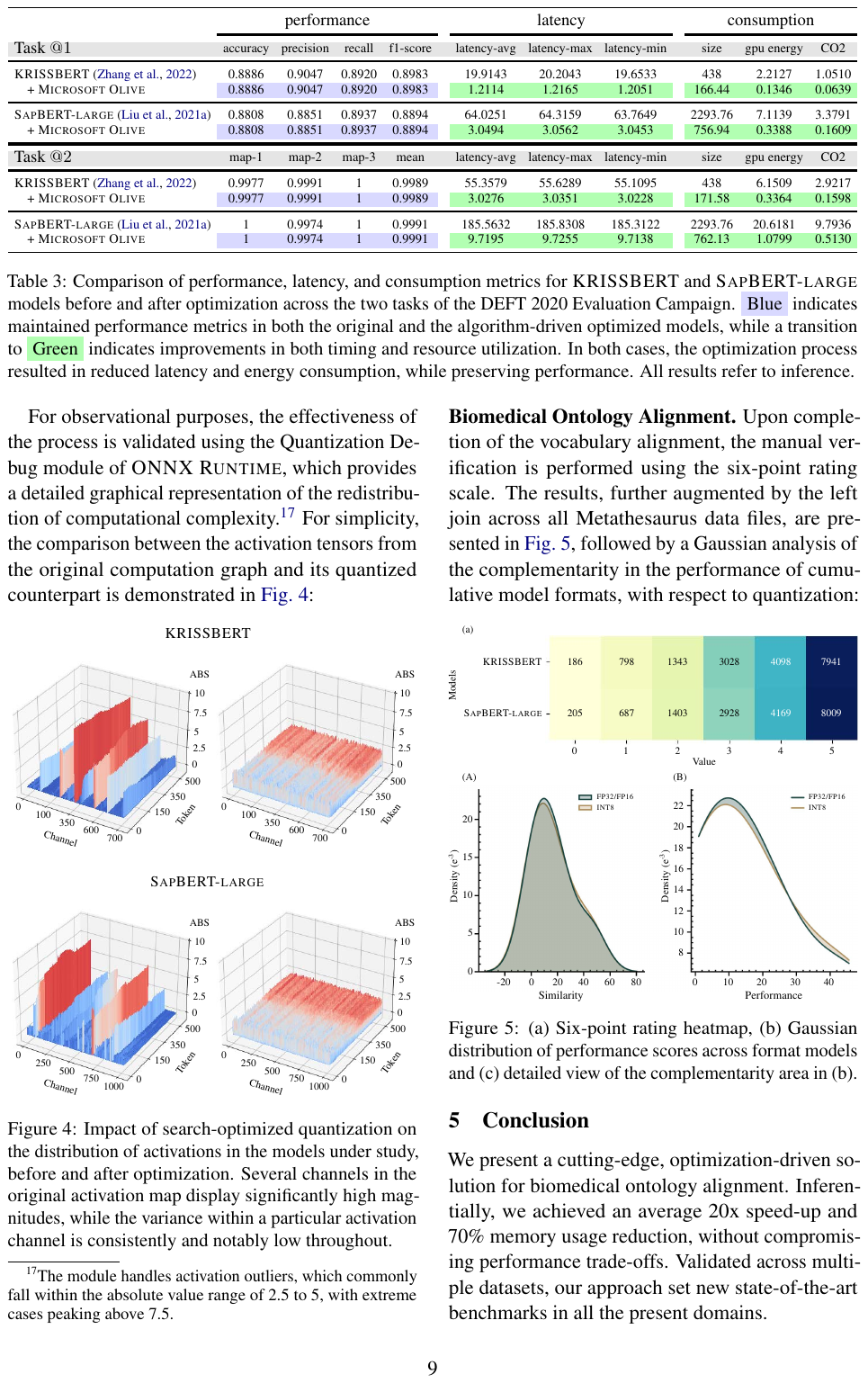}
    \vspace{-1.35em}
    \caption{(A) Gaussian kernel density estimation of performance scores across model formats; (B) Detailed view of distribution shifts induced by format variation.} 
    \label{fig:ontology_alignment_evaluation}
    \vspace{-0.3em}
\end{figure}

\section{Conclusion}
We present a cutting-edge, optimization-driven solution for biomedical ontology alignment. Inferentially, we achieved an average 20x speed-up and 70\% memory usage reduction, without compromising performance trade-offs. Validated across multiple datasets, our approach set new state-of-the-art benchmarks in all the present domains.
\section*{Limitations}
The performance of our methods is influenced by external factors, including hardware configurations, software dependencies, and environmental conditions. A thorough analysis of these elements and their impact is essential for practical deployment and real-world applications. Such analysis should also be extended to different model architectures, including large language models.
\section*{Acknowledgements}
We extend our sincere gratitude to the reviewers from the University of Lille and Centrale Lille for their insightful comments and suggestions, which significantly improved the quality and rigor of our work. We also thank CNRS UMR 8163 STL for providing the valuable resources and contributions essential to the success of this research.

\bibliography{naacl2021}
\bibliographystyle{acl_natbib}

\appendix
\clearpage

\appendix
\section{Evidence of the Analysis Error} \label{sec:proof}

\begin{table}[H]
\small
\centering
\vspace{-0.2em}
\begin{tabular}{@{}p{0.49\textwidth}@{}}
\toprule

\textbf{Source}: ``\emph{Royal jelly is a natural product very rich in vitamin B5 (C0001535), trace elements, acetylcholine (up to 0.1\% by mass), and antibiotic factors notably active against Proteus and Escherichia coli B (C0001041), better known as colibacillus.}''\\
\textbf{Target}: ``\emph{Indeed, the smoke (C0037369) makes the bees (C0005108) perceive a fire, causing them to frantically gather honey reserves in their crop rather than defending their hive from the beekeeper.}'' \\
\\
\textcolor{Maroon}{\textsc{KRISSBERT Prediction Score}: 95\%.} \\
\textcolor{NavyBlue}{{\ \ \ \ + \textsc{Corrective Fine-Tuning: 12\%.}}}\\
\\
\textcolor{Maroon}{\textsc{SapBERT-large Prediction Score}: 43\%.}\\ 
\textcolor{NavyBlue}{{\ \ \ \ + \textsc{Corrective Fine-Tuning: 7\%.}}}\\

\midrule

\textbf{Source}: ``\emph{The degrees of originality (C0006267) and hybridization (C0020155) of these breeds, as well as their homogeneity, are poorly described.}'' \\
\textbf{Target}: ``\emph{Without this precaution when opening a hive, the excitement of a colony can rise, making it very dangerous (C0205166), given the number of bees (C0005108).}'' \\
\\
\textcolor{Maroon}{\textsc{KRISSBERT Prediction Score}: 94\%.} \\
\textcolor{NavyBlue}{{\ \ \ \ + \textsc{Corrective Fine-Tuning: 9\%.}}}\\
\\
\textcolor{Maroon}{\textsc{SapBERT-large Prediction Score}: 37\%.}\\ 
\textcolor{NavyBlue}{{\ \ \ \ + \textsc{Corrective Fine-Tuning: 5\%.}}}\\

\bottomrule
\end{tabular}
\vspace{0.1em}
\caption{Examples highlighting a critical issue of score overestimation in the predictions made by the \textsc{KRISSBERT} and \textsc{SapBERT-large} models, which tend to disproportionately inflate the re-ranking scores, even for incomplete or incorrect entity matches.}
\label{tab:improvement}
\end{table}

\section{Fine-Tuning} \label{sec:fine-tuning} The fine-tuning configuration of the respective models involved defining architecturally optimal setups to ensure stability and effectiveness in tasks 1 and 2 of the DEFT 2020 Evaluation Campaign. The data preparation for the training modules associated with them adhered to the methodology reported for biomedical alignment within the main scope (\Cref{sec:preprocessing}), unifying task initialization into a cohesive and standardized approach.
In the Microsoft Research model, the Adam optimizer, in its \textsc{AdamW} variant, is employed with an initial learning rate of \(1 \times 10^{-5}\) and a learning rate scheduler, \texttt{ReduceLROnPlateau}, which reduces the learning rate by a factor of \(0.1\) if performance on the validation set does not improve over three consecutive epochs. The framework utilized is the \textsc{HuggingFace Trainer}, which streamlines the integration of model configuration, dataset preprocessing, evaluation metrics, and resource management within a unified execution pipeline. The \texttt{batch\_size} is set to \(8\), with a dropout rate of \(0.1\) applied to mitigate overfitting. Additionally, the \texttt{pmask} and \texttt{preplace} functions are implemented during tokenization with a probability of \(0.2\), thereby introducing controlled variability into the input data. The temperatures \(\tau\) and \(\pi\) are consistently maintained at \(1.0\) to ensure stable gradient flow.  
In the Cambridge LTL model, similarly to the Microsoft Research setup, the \textsc{AdamW} optimizer is applied with a learning rate of \(1 \times 10^{-5}\), but with a weight decay rate of \(1 \times 10^{-2}\). Although automatic mixed precision is originally preferred by researchers at Cambridge LTL, it is disabled in favor of maximum precision. The preprocessed and encoded data are partitioned into batches of \(8\) samples, across \(3\) epochs, with training likewise performed using the \textsc{HuggingFace Trainer}. Fine-tuning for both models is conducted on NVIDIA A100 GPUs, with all parameters carefully configured with respect to the nuanced connotative context specific to each of the two distinct and interrelated tasks.

In Task 1, the models undergo multi-class fine-tuning, utilizing a customized grading scale tailored to the scalably distorted cosine similarity outputs of the study models. This process involved converting the original labels from the \emph{t1-train} module into a percentage format of cosine similarity, scaled in accordance with the range of outputs obtained during an initial inference on \emph{t1-test}. This manipulation is necessary to properly test this technique statistically, allowing it to capture more nuances in the pairs of sentences of interest (\emph{source} and \emph{target}) compared to traditional binary class fine-tuning.
The loss function is adapted using a combined loss integrating categorical cross-entropy and mean squared error, also known as MSE. This choice is motivated by the fact that categorical cross-entropy loss is suitable for multi-class classification and allows the model to learn to correctly distinguish between different classes of semantic similarity. By incorporating mean squared error, predictions that deviate substantially from the actual similarity values are penalized, thus improving the model's accuracy in recognizing semantic gradation and its performance on the official evaluation metrics. The weights of the losses, $\alpha$ and $\beta$, are balanced at \(0.5\), ensuring harmonious optimization, in accordance with Eq.~\eqref{eq:combined-loss}:
\begin{equation}
\label{eq:combined-loss}
\alpha \times \text{Categorical Cross-Entropy} + \beta \times \text{MSE}
\end{equation}
Within this analytical framework, it is important to note that, prior to the optimization process, both modules (Train and Test) undergo a binary balancing between the positive and negative classes, the latter being slightly predominant. This is achieved through an automated undersampling method selectively applied to correct errors arising from discrepancies in human evaluation, notably when there is a significant distance between the \emph{mark} and \emph{mean} fields. An illustrative case is provided by the pair with identifier \emph{id}~\(413\) in the \emph{t1-train} module, where the \emph{mark} field has a value of \(5\), projectively corresponding to a positive label, yet the \emph{mean} field holds a projectively negative value of \(2.1\). The related \emph{scores} field is \([3, 0.5, 2, 5, 0]\), which logically should not yield a \emph{mark} value of \(5\), revealing an evaluation coherence error. Or, in the \emph{t1-test} module, by the pair with identifier \emph{id} \(38\), where there is a projectively positive value of \(4\) in the \emph{mark} field associated with the lower value of \(2\) in the \emph{mean} field, with the observed values \([2,1,0,3,4]\) in the \emph{scores} field.
This corrective adjustment did not significantly affect the original data composition, as both the distributions examined in each module remained quantitatively similar. Nevertheless, it contributed to more reliable and representative evaluation criteria, mitigating instability introduced by inconsistent assessments.

In Task 2, the models are trained with the aim of improving the identification of correspondences between pairs of sentences of interest (\emph{source} and \emph{target}) through the calculation of cosine similarity. This objective is pursued by adhering to the underlying logic of simple-complex relationships in sentence parallelism, taking into account three distinct conditions within each compartment. This compartmental structure is aligned with the purpose of the task, namely to evaluate three candidate \emph{target} sentences and determine the one that exhibits the highest degree of parallelism with the \emph{source} field. Given that a response is always expected from the three provided \emph{target} sentences, the task requires the identification of a suitable parallel sentence for each corresponding set of \emph{source} and \emph{target} sentences. In reconsideration, the concept of sentence parallelism is rooted in the simple-complex relationship, wherein the \emph{source} sentence represents complex content, while the simple sentences convey simplified or less complex content, resulting from derivation.
A list of positive pairs (\emph{source} and \emph{target}), sequentially initialized with \texttt{[CLS]} tokens and then concatenated and delimited with \texttt{[SEP]} separator tokens, is generated by combining the respective correspondences with \emph{target}. This sequence is then passed through a higher classification layer, which identifies the correct alignment via the correspondence with \emph{num} in each tripartite compartment for all unique identifiers \emph{id}.
In this context, the loss function is simplified in comparison to the previous one, as it is based on the cross-entropy loss. This allows the models to learn to accurately minimize the loss by directly comparing the predictions with the true labels (\emph{target}). These parametric finalizations involve prototyping a series of trial calibrations in the respective Test phases, thereby determining the optimal values according to the functional properties of each model.

\vspace{3.61pt}

\section{Why is it important to apply the sentence-similarity modality?}
Upon in-depth consideration, opting for a conventional \texttt{text-classification} task would have resulted in an evaluation metric not suitable for our study, as the six-point similarity scale employed by the five expert annotators of the DEFT 2020 Evaluation Campaign is explicitly designed to assess contextual cosine semantic similarity. Therefore, quantifying each sample using cosine similarity and subsequently adapting the inference output distribution to match the official multi-label evaluation format proves to be the most appropriate approach. For methodological purposes, the task is framed in \texttt{sentence-similarity} mode to demonstrate the benefits of optimization, specifically maintaining performance metrics while simultaneously reducing latency and resource consumption. This is carried out within an experimental setting that is intrinsically aligned with both the biomedical focus of our core objective and the DEFT 2020 evaluation framework.

\vspace{3.61pt}

\section{License of Scientific Artifacts}
\label{sec:license}

UMLS \citep{bodenreider2004unified} is licensed to individuals for research purposes. CNRS resources are provided under the End User License Agreement (EULA), as are the DEFT 2020 Evaluation Campaign datasets \citep{cardon-etal-2020-presentation}. 
The MedSTS dataset \citep{journals/corr/abs-1808-09397} is freely available for public use.
\textsc{KRISSBERT} \citep{zhang2022knowledgerichselfsupervisionbiomedicalentity} and \textsc{SapBERT-large} \citep{liu2021selfalignmentpretrainingbiomedicalentity} models are distributed under the MIT License, as are \textsc{Microsoft Olive} and \textsc{ONNX Runtime}.
\textsc{ScispaCy} \citep{neumann-etal-2019-scispacy}, \textsc{Intel Neural Compressor}, and \textsc{IPEX} (Intel Extension for PyTorch) are released under the Apache License 2.0.

\label{proof}


\end{document}